\documentclass{article}  

\usepackage[utf8x]{inputenc}
\usepackage[english]{babel}
\usepackage{fancyhdr}
\usepackage[top=2cm,bottom=2cm,left=2cm,right=2cm,marginparwidth=1.75cm]{geometry}
\usepackage{graphicx}
\usepackage{caption}
\usepackage{subcaption}
\usepackage{float}
\usepackage{setspace}
\usepackage{soul}
\usepackage{pstricks}
\usepackage{pst-circ}
\usepackage{hyperref}
\usepackage{dblfloatfix}
\usepackage{amssymb}
\usepackage{gensymb}
\usepackage{amsmath}
\usepackage{dblfloatfix} 
\usepackage{multicol}

\title{\LARGE \bf Hold 'em and Fold 'em:
Towards Human-scale, Feedback-Controlled Soft Origami Robots\footnote{This work is supported by the National Science Foundation under Award ECCS 1846954}}
\author{Immanuel Ampomah Mensah, Jessica Healey, Celina Wu \\  Andrea Lacunza, Nathaniel Hanson, and Kristen L. Dorsey\\ Institute for Experiential Robotics, Northeastern University, Boston, USA}
\date{}

\begin{document}

\maketitle

\begin{abstract}
%Please include an overview of your project. This must include at least one sentence sharing the significance of your project in the soft robotics field. 

An underdeveloped capability in soft robotics is proprioceptive feedback control, where soft actuators can be sensed and controlled using only sensors on the robot's body. 
Additionally, soft actuators are often unable to support human-scale loads due to the extremely compliant materials in use. 
Developing both feedback control and the ability to actuate under large loads (e.g. 500 N) are key capacities required to move soft robotics into everyday applications. 

In this work, we independently demonstrate these key factors towards controlling and actuating human-scale loads: proprioceptive (embodied) feedback control of a soft, pneumatically-actuated origami robot; and actuation of these origami origami robots under a person's weight in an open-loop configuration.
In both demonstrations, the actuators are controlled by internal fluidic pressure. 
Capacitive sensors patterned onto the robot provide position estimation and serve as input to a feedback controller.   
We demonstrate position control of a single actuator during stepped setpoints and sinusoidal trajectory following, with root mean square error (RMSE) below 4 mm. 
We also showcase the actuator's potential towards human-scale robotics as an ``origami balance board'' by joining three actuators into an open-loop controlled system with a platform that varies its height, roll, and pitch. 
\emph{This work contributes to the field of soft robotics by demonstrating closed-loop feedback position control without visual tracking as an input and lightweight, soft actuators that can support a person's weight}. The project repository, including videos, CAD files, and ROS code, is available at \url{https://parses-lab.github.io/kresling_control/}

\end{abstract}

\section{Introduction and Background Knowledge}
%Are there any relevant principles, information or research someone should be aware of in order to replicate your project? (~1 paragraph

\begin{figure}[h]
    \centering
    \includegraphics[width=6.5in]{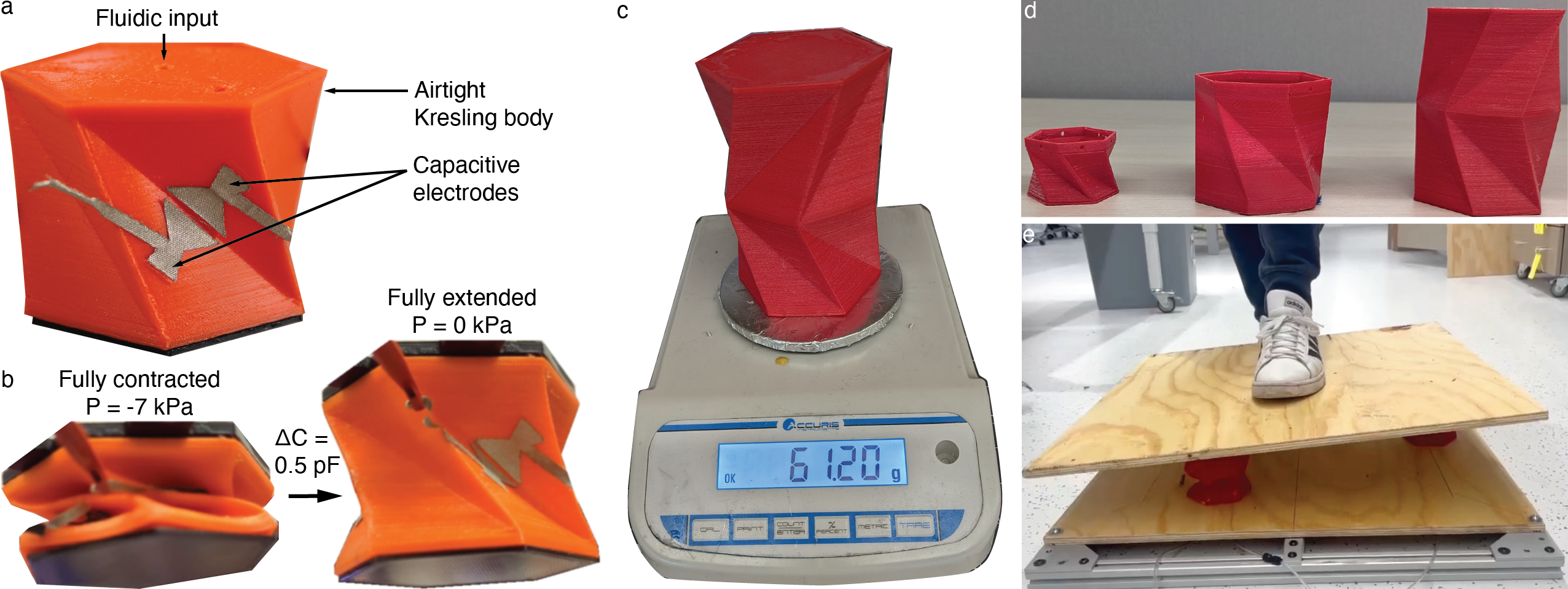}
    \caption{The origami actuator (a) overview and (b) under fluidic actuator. Photographs of (c) the low actuator mass, (d) actuators at different scales, and (e) a rider preparing to step onto the demonstration balance board.}
    \label{fig:teaser}
\end{figure}

Soft robots may soon be part of daily life, with envisioned applications ranging from helping humans \cite{robertson2021soft, wu_stretchable_2021} to holding fragile objects without causing damage \cite{li_vacuum-driven_2019}.
Reliably controlled, precise, and soft actuators are a critical part of this vision.
Within the wide array of soft actuator designs and approaches, soft origami actuators fabricated from paper or polymers offer mechanical compliance and have the advantage of programmable deformation through a pattern's folds \cite{park_reconfigurable_2022}.
Insight gained from early work in origami robot fabrication, actuation, and sensing approaches \cite{park_reconfigurable_2022,chen_origami_2022, robertson2021soft} spurred interest in more complex patterns and motions. 
One common design for origami soft robots is the Kresling pattern \cite{kresling_origami-structures_2012, kim_3d_2022, murali_babu_programmable_2023}, which is a cylindrical origami shape that is representative of the buckling pattern formed when a cylinder of paper is twisted and compressed. 
While the Kresling pattern originated as a paper structure, it can be 3D printed out of other compliant materials such as thermoplastic polyurethane (TPU). 
3D printing enables Kresling modules to be fabricated with consistency and ease, and simplifies the process of adjusting its dimensions.
Prior work has applied the Kresling pattern to create crawlers \cite{ze_soft_2022} and multi-segment continuum arms \cite{kaufmann_harnessing_2021} under magnetic, cable-driven, or fluidic \cite{murali_babu_programmable_2023} actuation.

In contrast to the wealth of soft mechanics and actuator work present in origami robotics literature, position sensing and feedback control research has received less attention.
A critical capability will be accurate state estimation, position control, and force control \cite{best_new_2016, tapia_makesense_2020} that will be essential for integrating soft robots into daily life.
While camera-based tracking \cite{della_santina_dynamic_2018, patterson_untethered_2020} is one approach for position or force sensing, it is not well suited to portable or high-privacy applications. 
Soft mechanical sensors (e.g., capacitors \cite{kim20213d}, resistors \cite{sun_repeated_2022}, or inductors \cite{wang_stick_2022}) affixed to the robot body can provide accurate force and shape estimation without cameras \cite{yan_origami-based_2023, sun_origami-inspired_2022}.
With this submission, we present a feedback-controlled one DoF origami-inspired robot that builds upon our previous findings \cite{hanson_controlling_2023}. 
The position of the robot is determined by capacitive sensors adhered to the actuators (Fig. \ref{fig:teaser}a).
We chose capacitance as the sensing modality for its high sensitivity, potential for integration into soft fabrication processes, and ability to measure with compact electronics. 
We use fluidic actuation to drive the robot contraction (Fig. \ref{fig:teaser}b). Our approach offers greater portability and mechanical simplicity as the design contains no internal components. 

To demonstrate the potential of this actuator in real-world systems, we assembled three actuators into an open-loop controlled ``origami balance board.'' Three actuators with total mass less than 200 g (Fig. \ref{fig:teaser}c) can support the mass of a 157 cm tall rider (Fig. \ref{fig:teaser}e).

This work contributes to the state-of-the-art in soft robotics proprioceptive control by:
\begin{enumerate}
\vspace{-0.5em}
\item designing and optimizing capacitive position sensors using finite element analysis (FEA);
\vspace{-0.5em}
\item programmatically-generating and fabricating Kresling robots of different scales through 3D printing; and 
\vspace{-0.5em}
\item demonstrating feedback control using only the capacitive electrodes. 
\end{enumerate}

\section{Design Component Overview}
%An overview of the different components of your project. What materials and tools would someone need in order to replicate your project fully?

\begin{figure}[h]
    \centering
    \includegraphics[width=2.5in]{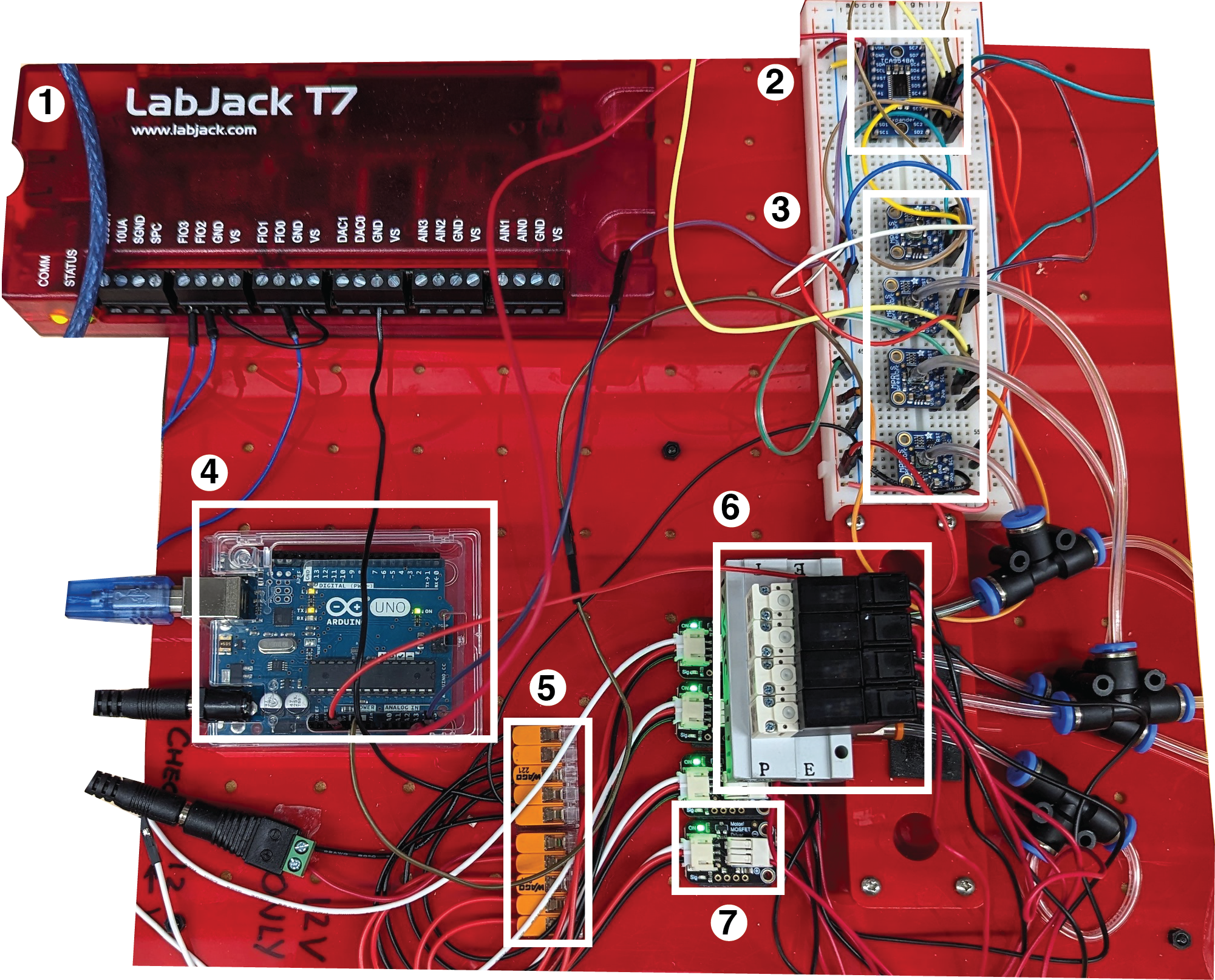}
    \includegraphics[width=2.5in]{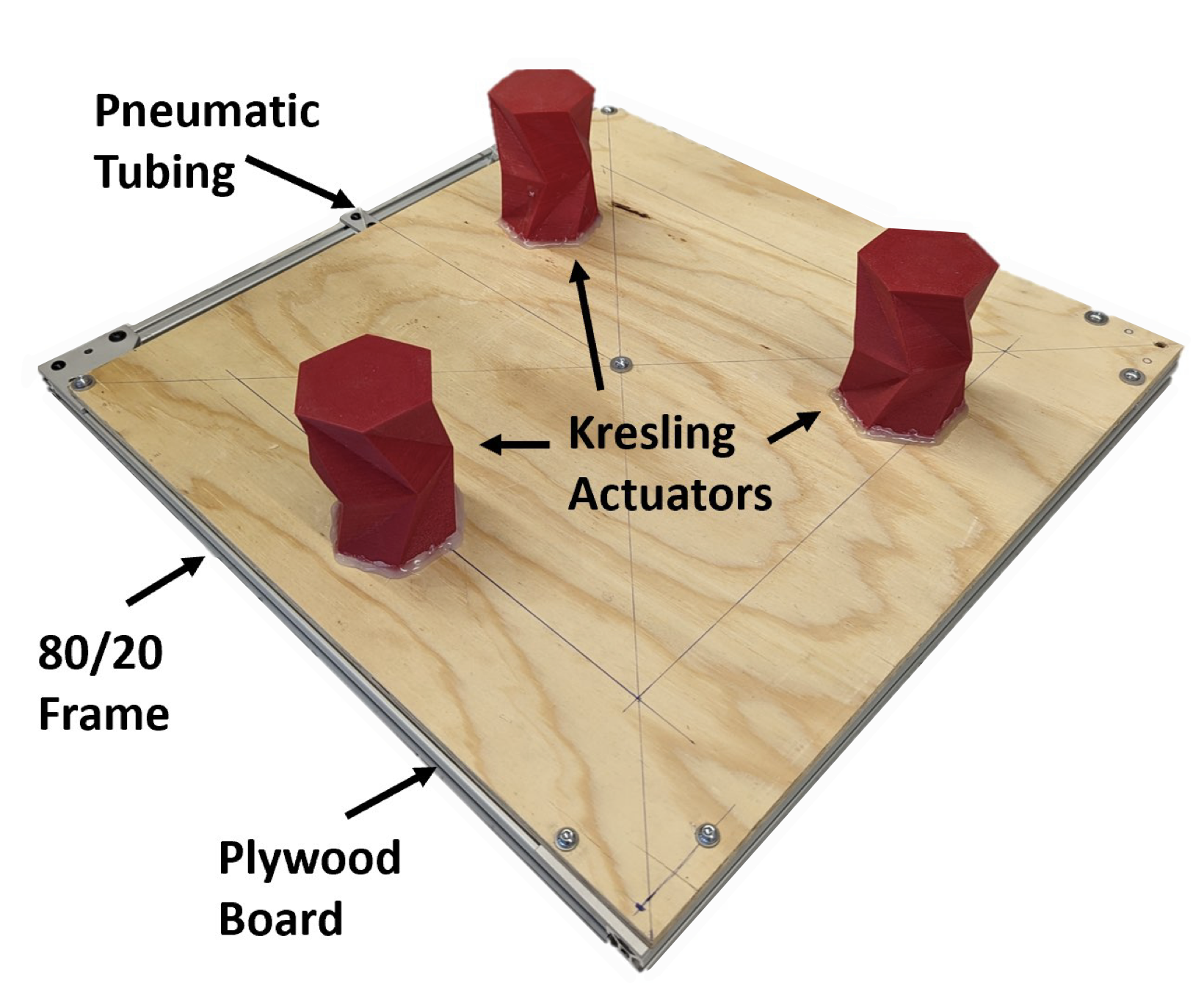}
    \caption{(left) The fluidic control board, further discussed in section 6.2
    (right) A photograph of the system components including: Kresling actuators, the base plate, and pneumatic tubing entering the side of the base frame.}
    \label{fig:board}
\end{figure}

The components of this project include Kresling actuators, capacitive sensors, and a fluidic control board (Fig. 2). The case study combines three actuators with two rigid plates to form a balance board with an actuator mass of 182 g and a total board mass of 8.6 kg for the frame, plates, and actuators. A pressurized air source is fed into the fluidic control board and separated into a pneumatic tube for each Kresling actuator. Pneumatic tubing runs from the fluidic control board to the base of the Kreslings. The first rigid plate acts as a base plate for the bottom of the Kreslings. The second rigid plate rests on top of the actuators and serves as the platform for a rider. Materials required to fabricate the frame design are pneumatic tubing, glues, sealants, and rigid boards (e.g., plywood sheets) to act as base plates.

\section{Material Selection}
%How were the materials decided on for the project? Are there alternate material solutions that would be adequate, but simplify the design, cost, etc.?

\begin{figure}[h]
    \centering
    \includegraphics[width=2 in]{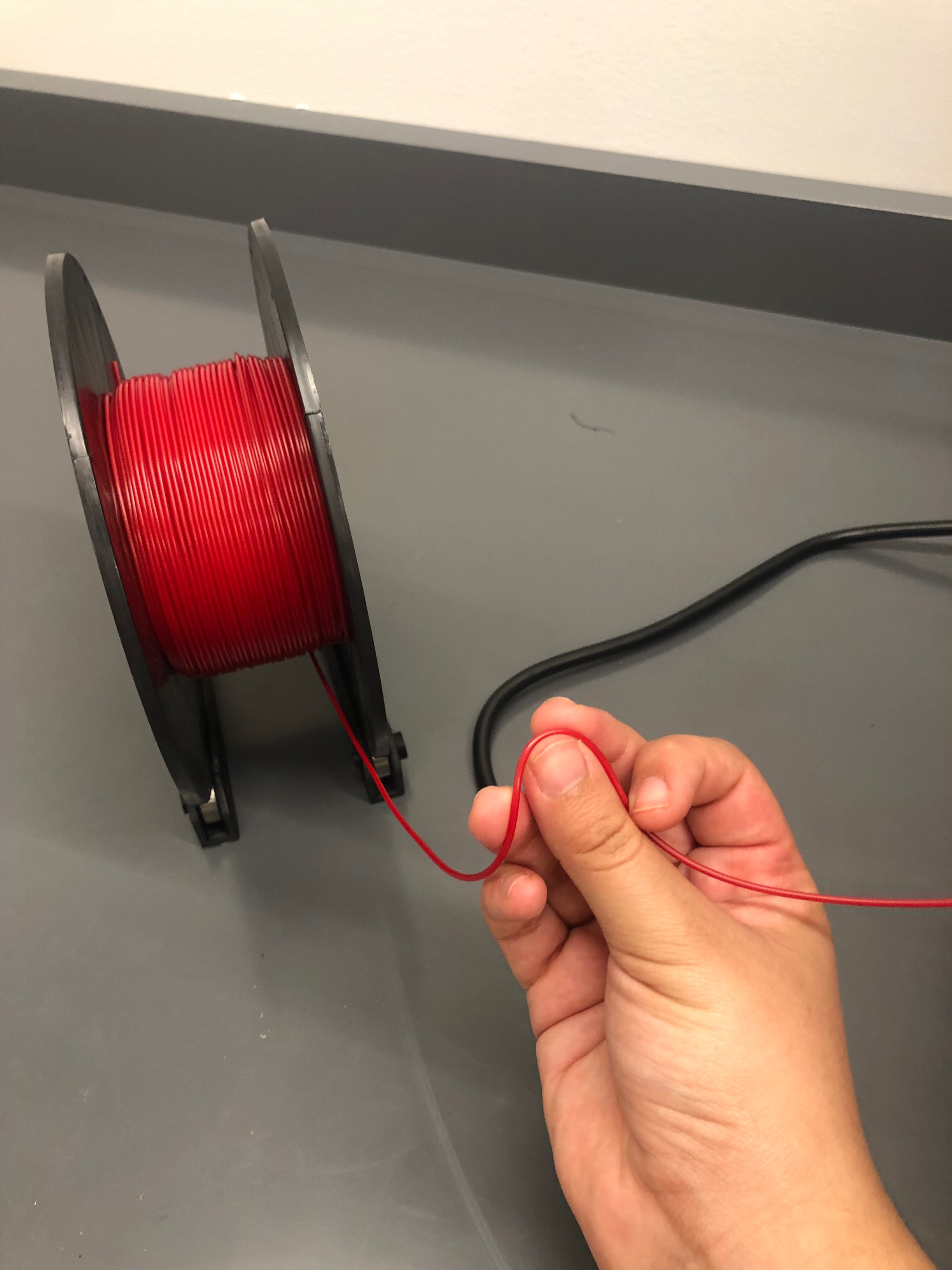}
    \caption{A photograph of the flexible TPU filament used for 3D printing.}
    \label{fig:TPU}
\end{figure}

To fabricate our Kresling actuators, we utilized Thermoplastic Polyurethane (TPU) filament (Ninjaflex Cheetah, Ninjatek)(Fig. \ref{fig:TPU}) in conjunction with a 3D printer due to its flexible material properties and cost-effectiveness. The flexibility and conformability of TPU render it an ideal choice for constructing dynamic origami structures. 
It is also non-porous and airtight, which permits actuation with negative and positive fluidic pressure. 

One disadvantage of TPU is that it is highly hygroscopic, so print quality degrades with increased humidity. As a result, we periodically dry the filament using a commercial filament dryer to ensure optimal print quality. While alternative materials such as polymers, paper-like substances, or thin metals could have been considered for manufacturing, they would have entailed longer lead times, less control over the structure angles, and potentially higher costs.

Copper and nickel-plated polyester fabric from Adafruit were selected for the electrodes due to their flexibility and electrical conductivity. 
The fabric's material structure allowed for easy placement between the folds of the Kresling origami actuator. Conductive fabric also exhibits excellent mechanical resistance to torsion and tearing, rendering it suitable for dynamic operations. Several alternatives to copper and nickel-plated polyester fabric were considered, including conductive polymers, elastomers, and fabrics with different metal compositions. However, these alternatives either do not offer high conductivity, exhibit greater weight and reduced flexibility, or delaminated from the robot during actuation. 

\section{Design Optimization}
%If experiments were performed to optimize the design, what went into this design process? 

\subsection{Origami Actuator Structure and Kinematic Model}
\begin{figure}[h]
    \centering
    \includegraphics[width=3.25in]{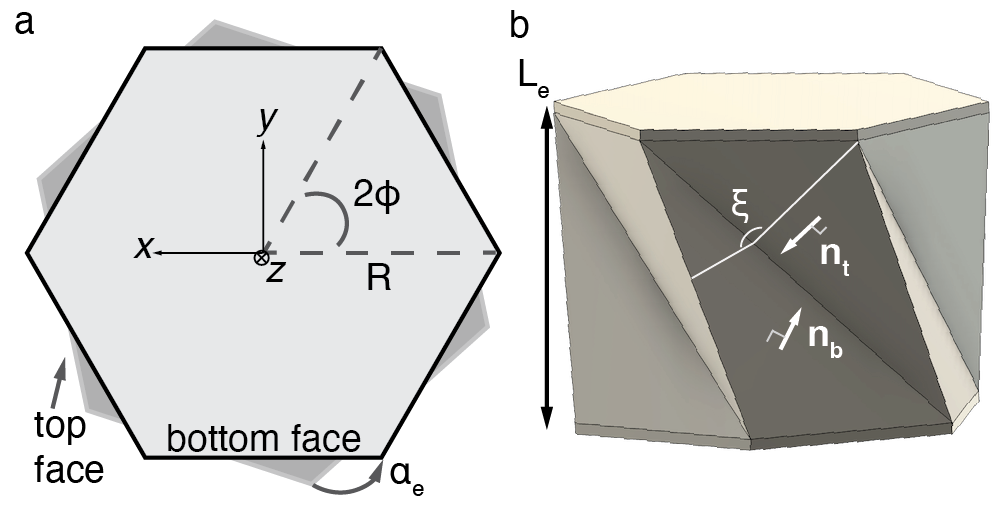}
    \caption{The parameters of the Kresling structure (a) from the bottom face looking up and (b) in 3D}
    \label{fig:parameters}
\end{figure}

Previous work has thoroughly investigated the mechanics of the Kresling pattern \cite{kaufmann_harnessing_2021, bhovad_peristaltic_2019}. %
Here, we present a short definition of properties relevant to the modeling of the Kresling fold. %
The structure has a single DOF, as described by linked parameters $L$, which is the distance between the top and bottom faces, and $\alpha$, which is the rotation between the faces.
Four parameters control the morphology: the distance between top and bottom faces at full contraction $L_c$, the circumscribing radius around the top face $R$, the angle ratio $\lambda$ between the valley fold and a bottom edge to the angle between a bottom edge and a line that bisects a bottom face inner angle, and the number of Kresling polygon edges $N$. 
The interior angle formed between the center and one edge of the top face is $\phi = \pi/N$. %
Fig. \ref{fig:parameters} represents these parameters on the Kresling structure with views (a) in 2D, looking up through the bottom face, and (b) in 3D.

These variables are linked to one another \cite{kaufmann_harnessing_2021} through 
\begin{equation}
    L = \sqrt{L_c^2 +2R^2\left[ \cos(\alpha + 2 \phi) - \cos (\alpha_c + 2\phi)\right]} ,
    \label{eqn:length}
\end{equation}
where $L$ and $\alpha$ represent a distance and rotation, respectively, between the top and bottom faces, and $\alpha_c$ = 2$\lambda(\pi/2-\phi)$ is the rotation at full contraction. 

We will refer to three additional dependent parameters to describe the Kresling structure.
At full extension, the structure has length $L_e$ and rotation $\alpha_e$.
Two triangular sidewall faces, one with an edge on the top and the other with an edge on the bottom polygon face, intersect to form a valley fold with dihedral angle $\xi$. %

The fold angle $\xi$ is linked to $L$ by finding the angle of intersection between vectors normal to the sidewall triangular faces $n_t$=
\begin{equation}  
\begin{bmatrix}
RL \sin(2\phi) \hat i\\
RL \left( 1 - \cos(2\phi) \right) \hat j\\
R^2 \left( \sin(\alpha)(1-\cos(2 \phi))+(1-\cos(\alpha))(\sin(2 \phi)) \right) \hat k\\
\end{bmatrix}
\end{equation}

and  $n_b$=
{
$\displaystyle$
\begin{equation}
\begin{bmatrix}
RL (\sin(2\phi - \alpha) + \sin(\alpha)) \hat i\\
-RL (\cos(2\phi - \alpha) - \cos(\alpha)) \hat j\\
R^2 \left( (\cos(2\phi - \alpha)-\cos(\alpha))(\sin(2\phi - \alpha)+\sin(2\phi)) \right) \hat k -\\
R^2\left( (\cos(2\phi - \alpha)-\cos(2\phi))(\sin(2\phi - \alpha)+\sin(\alpha)) \right) \hat k\\
\end{bmatrix}
\label{eqn:nvecs}
\end{equation}
}
where $\hat i$, $\hat j$, and $\hat k$ are unit vectors on the axes $x$, $y$, and $z$, respectively, such that 

\begin{equation}
   \cos \xi = \frac{\vec n_t \cdot \vec n_b}{||n_t||||n_b||} . 
   \label{eqn:xi}
\end{equation}
    
We derived the relationship between $L$ and $\xi$ through Eqs. \ref{eqn:length}$-$\ref{eqn:xi}.
Observing that the relationship between fold angle and length is highly linear via a numerical approach ($R^2$ of $99.3\%$ in the range of $L_c$ to $L_e$), we used a linear approximation for the inverse kinematics function $L = 0.22\xi + 10.4$ mm, where $\xi$ is in degrees.

\subsection{Capacitive Sensor Optimization}
Capacitive sensing offers high sensitivity and ease of readout with measurement electronics, making it suitable for displacement sensing in soft robots. 
One potential disadvantage of sensing the pose of the origami robot with capacitors is the relationship between fold angle, electrode distance, and capacitance.
Although the relationship between $\xi$ and $L$ is well approximated with a linear function, the reciprocal relationship between electrode distance and capacitance will yield much lower capacitance at large (more obtuse) fold angles. 
In a Kresling structure where nearly the entire sidewall face is covered with an electrode, these super-linear behaviors will yield a low sensitivity to displacement when the robot is nearly extended.

To obtain a higher sensitivity at larger fold angles, we optimized the electrode shape. 
Due to the lack of closed-form solutions for various electrode shapes, we used FEA (COMSOL v6.0) and a downhill simplex method implemented in MATLAB 
 \cite{lagarias1998convergence} (\texttt{fminsearch}) for optimization.
This approach does not rely on an analytical solution of the gradient (e.g. Newton's method) and was computationally feasible with FEA (e.g., vs. genetic algorithm).
Electrode shapes were represented by five values that control the horizontal and vertical placement of points on the electrode face.
The optimization function was:
\begin{gather}
\text{maximize }f(x) = \frac{\Delta C_{65}(x)}{C_{10}(x)}\end{gather}
where $\Delta C_{65}(x)$ is the capacitance change of a fold moving between 70$\degree$ and 65$\degree$ from a horizontal plane (i.e., $\xi$/2) and is constructed of the vertices in the subset $x$, which is a subset of all potential electrode points in $\mathbb{R}^2$ bounded by the sidewall face. $C_{10}(x)$ is the capacitance at 10$\degree$ also constructed of the same subset vertices $x$.

\begin{figure}[h]
    \centering
    \includegraphics[width=3.25in]{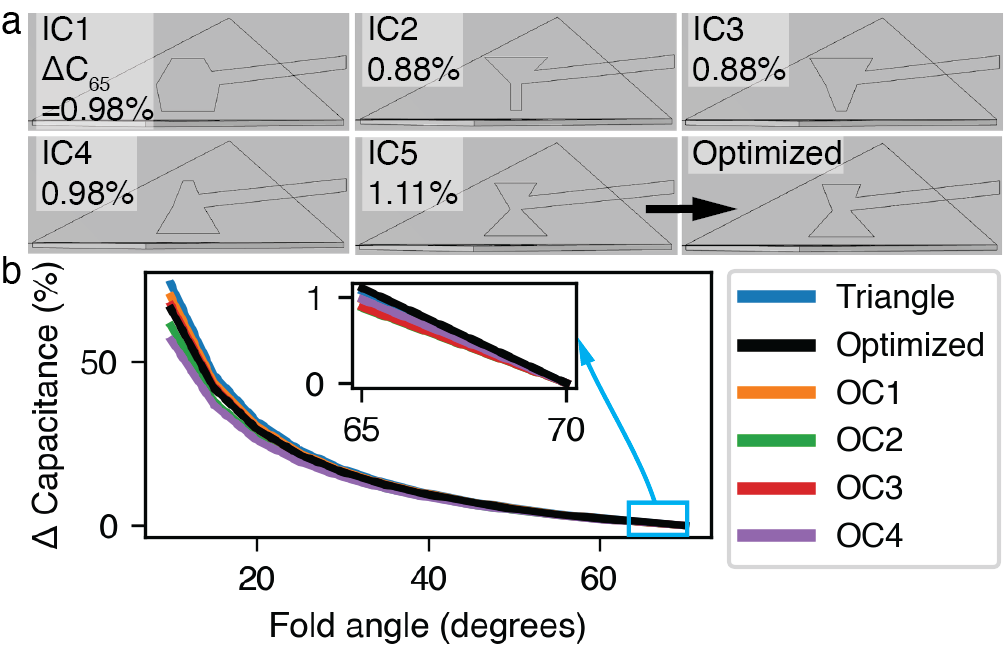}
    \caption{Finite element electrode modeling. (a) The initial electrode shapes IC1-IC5 and OC5. The optimized sensitivities are presented in the top left corners.  (b) The simulation results for electrode shapes OC1$-$OC5 and the triangle electrode. \emph{Inset:}  $\Delta C_{65}$.}
    \label{fig:FEA}
\end{figure}

We chose five initial conditions (IC) for the electrode shapes (IC1$-$5) selected from a larger set for variations in initial capacitance and point placement to optimize our electrode shapes (Fig. \ref{fig:FEA}a).
From the optimized electrode (OC) from each initial condition, we selected the electrode with the largest change in capacitance at large fold angle for further testing (i.e., OC5). 
Each minimum fractional sensitivity is presented in Fig. \ref{fig:FEA}a, and the fractional capacitance change ($C-C_{70})/C_{10}$ is presented in Fig. \ref{fig:FEA}b.
The performance of a triangular electrode inset 1.125 mm from the valley fold ($\Delta C_{65}/C_{10}$ of 1.07\%) is also shown. 
Capacitance was also simulated for OC5 across the full range of $\xi$ from extension to contraction.

\section{Fabrication}

\begin{figure}[h]
\centering
\includegraphics[width=4in]{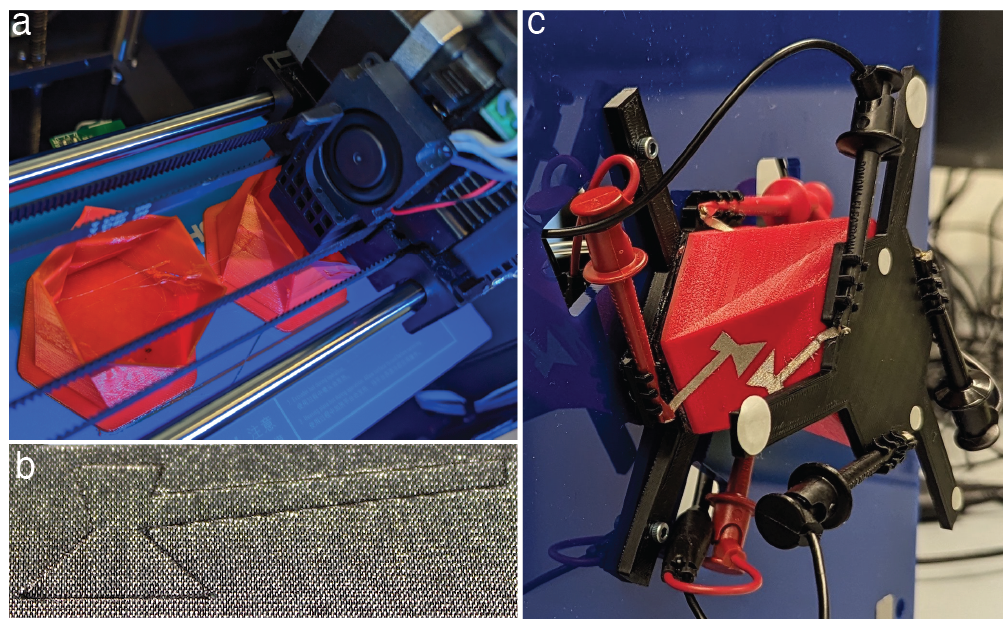}
\caption{The fabrication process. (a) 3D printing, (b) the electrodes after cutting, (c) the assembled robot with tracking markers for optical motion capture and ground truth}
\label{fig:fabrication}
\end{figure}

\subsection{3D Printing}
%Ash and Andrea-- nice photos, description of process
We 3D printed (Creator Pro 2, Flash Forge) the Kresling structure with TPU filament (NinjaFlex Cheetah 95A) (Fig. \ref{fig:fabrication}a). 
Print settings were a 0.4 mm nozzle at an extruder temperature of 238 \degree C, platform temperature of 50 \degree C, layer height of 0.18 mm, and print speed of 30 mms$^{-1}$. 

The STLs were sliced in Simplify3D. The detailed settings are listed in Table \ref{tab:slicing_settings}.
\begin{table}[h]
    \centering
    \begin{tabular}{ll}
         Setting& Value\\
         \hline
         Nozzle Diameter& 0.4mm\\
         Extrusion Multiplier& 1.09\\
         Use Retraction& Disabled\\
         Layer Height& 0.18mm\\
         Print Speed& 30mm/s\\
         External Thin Wall Type& Allow Single Extrusions\\
         Internal Thin Wall Type& Allow Single Extrusions\\
         Extruder Temperature& 238 \degree C\\
         Bed Temperature& 50 \degree C\\
    \end{tabular}
    \caption{Simplify3D Slicing Settings}
    \label{tab:slicing_settings}
\end{table}

Prior to initiating each print, the 3D Printer goes through a preheating procedure where the 3D printer's extruder and print bed are heated to their specified operating temperatures. 
Fabricating each human-weight Kresling actuators requires over nine hours for structure and lid, while the smaller Kresling structure requires three hours.

\subsection{Sensorization}
%Jess
To form the electrodes, we cut copper and nickel plated polyester fabric (Adafruit) into the electrode shape with an electronic cutting machine (Maker 3, Cricut) (Fig. \ref{fig:fabrication}b). %
We adhered these electrodes to opposite faces of the Kresling valley fold with double-sided pressure sensitive adhesive to form a capacitor. Three capacitors were symmetrically spaced radially around the $z$ axis. 
The placement of the electrodes was guided by extruded lines on the sidewalls.

\subsection{Sensorized Robot Assembly}
%Jess
To actuate and test the robot, we capped the structure with a TPU lid and cyanoacrylate glue (Loctite 495) and inserted pneumatic tubing through holes in the cap. %
Rigid plastic bases were bonded to the top and bottom faces of the Kresling actuator to enforce rigid bodies for kinematics and optical tracking. %
Fig. \ref{fig:fabrication}c is a photograph of the robot mounted to a motion capture benchmarking stand.

\subsection{Balance Board Assembly}

Three Kreslings were adhered radially around the center of a ridig plate using hot melt adhesive (Fig. \ref{fig:board}). 
Our design utilized plywood boards for base plates, and an 80/20 frame to provide clearance for the pneumatic tubing to enter the bottom of the Kresling actuators. 
Plywood and 80/20 are common and robust fabrication materials, however the base can be made out of any rigid material that leaves space for pneumatic tubes to enter the frame. 
An all-wood base would reduce the weight and cost of the system. 
We also used cyanoacrylate glue (Loctite 495) and epoxy (Xtc-3D High Performance 3D Print Coating) for joining the lid to the structure sealing the Kresling actuators. Any glue or sealant compatible with TPU can be used for this purpose.

\section{Design Process}
%What are the step by step instructions someone would need in order to replicate your project? (Please include photos/videos where possible)

\subsection{CAD Models}
%Ash-- nice screenshots and description of how the model works
To fabricate the Kresling robots, we programmatically generated a CAD model using a custom  script that runs on Autodesk Fusion 360’s Python API.
The script generates 3D Kresling models based on a set of adjustable user-input parameters. 
In this script, each fold of the actuator is drawn as two sets of lofted triangles.

To generate the CAD model using the script available through our GitHub repository:
\begin{enumerate}
    \item The Kresling.py script must be downloaded from the repository. Place the downloaded .py file into a folder titled ``Kresling.''
    \item Autodesk Fusion 360 must be downloaded from the Autodesk site. Autodesk permits educational licenses for Fusion 360 to be used for university research.
    \item Open Fusion 360 and select ``UTILITIES'' in the toolbar.
    \item Open the ``ADD-INS'' menu and select ``Scripts and Add-Ins''.
    \item Click the green ``+'' next to ``My Scripts'' in the ``Scripts'' tab.
    \item Navigate to the ``Kresling'' folder where the .py script lives.
    \item Hit ``Select Folder'' to add the script to Fusion 360.
    \item To run the script, select ``Kresling'' in the ``My Scripts'' list and click the ``Run'' button.
    \item When the script runs, a new document opens in Fusion 360. A window with a set of adjustable parameters allows the user to modify critical dimensions on the Kresling. The Kresling model automatically updates while the parameters are altered.
    \item To generate the Kresling with the selected parameters, hit ``OK''.
\end{enumerate}
\begin{figure}[h]
    \centering
    \includegraphics[width=3in]{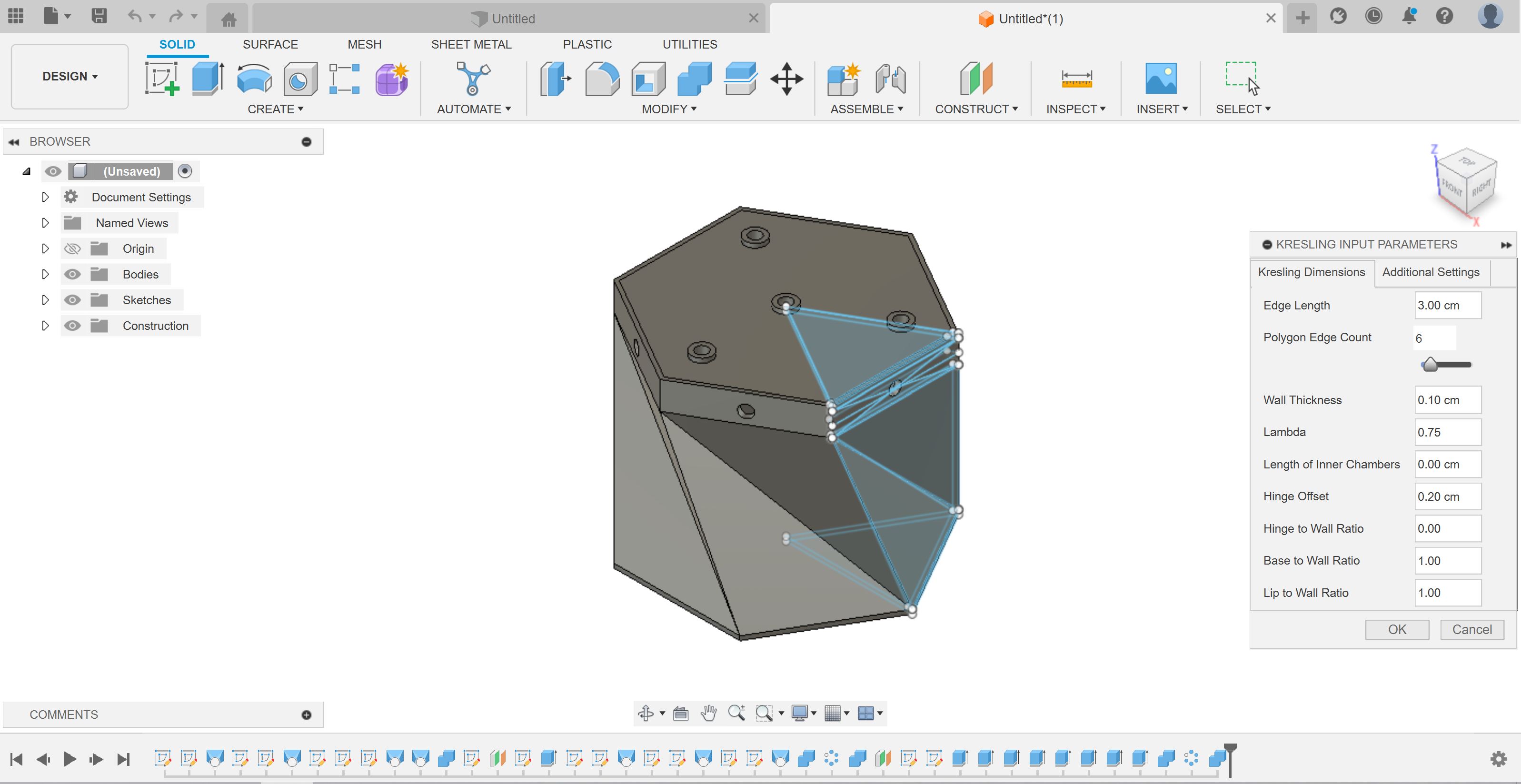}
    \caption{The parameters of the generated Kresling can be adjusted in Fusion 360}
    \label{fig:script_parameters}
\end{figure}

Once the CAD model is generated, these Kresling building blocks can be further modified in the CAD and altered based on the intended application. When the object is ready to be 3D printed, it can be exported as an STL:
\begin{enumerate}
    \item Open the ``Bodies'' dropdown on the feature tree.
    \item The model is separated into different bodies so each part can be printed separately (such as the lid and the main Kresling structure). The main Kresling structure can be identified by its name, which lists the values of key parameters used to generate the model.
\begin{figure}[h]
    \centering
    \includegraphics[width=3.25in]{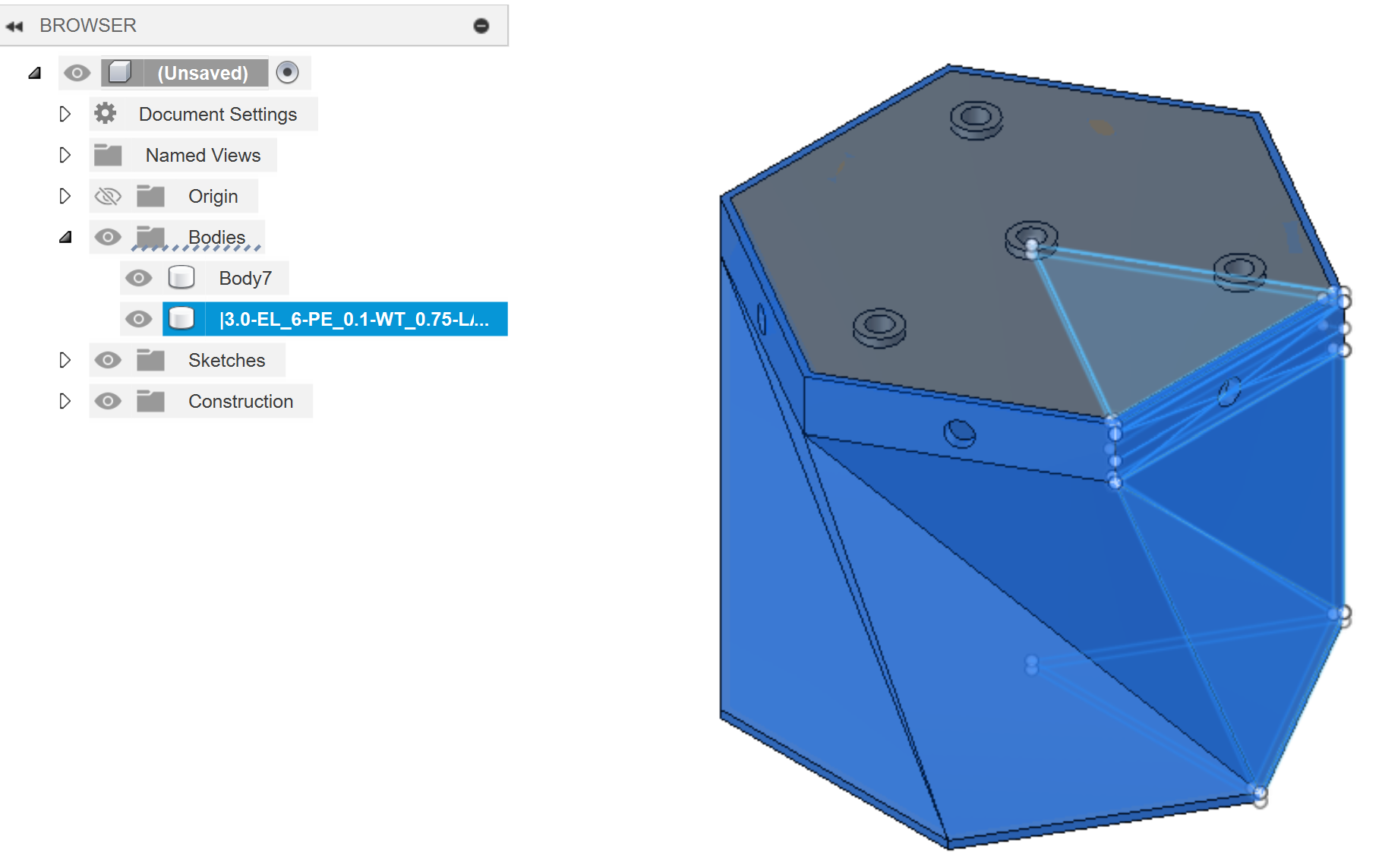}
    \caption{The main Kresling structure is named after the primary parameters that define its shape}
    \label{fig:kresling_name}
\end{figure}
    \item Right-click one of the bodies to export it as an STL. Select ``Save As Mesh.''
    \item Ensure the ``Format'' is set to ``STL (Binary)''.
    \item Hit ``OK.'' Give the STL an appropriate file name and click ``Save.''
\end{enumerate}

\subsection{Fluidic Control Board}
Actuator position is driven by a custom fluidic control board (Fig. \ref{fig:board}).
Fig. \ref{fig:board} is a photo of the custom fluidic control board. A data acquision system (T7, LabJack, \#1) sources PWM signals to the solenoids (VQ110U-6M, Orange Coast Pneumatics, \#6) to drive actuator pressure. The PWM signals at 5 V are converted to the 12 V solenoid voltage through a MOSFET driver (5648, Adafruit, \#7), and 12 V power is supplied with a switching power converter (352, Adafruit, not shown). Ground and 12 V signals are distributed via Wago snap-action connectors (784, Adafruit, \#5). An Arduino Uno (\#4), I2C multiplexer (2717, Adafruit, \#2), and ported absolute pressure sensors connected to the actuator working pressure (3965, Adafruit, \#3) complete the pressure measurement circuitry. An I2C multiplexer is required due to hard-coded I2C addresses on the pressure sensors.
Capacitance is read at a rate of 90 Hz with a multi-channel capacitance-to-digital converter (FDC2214 EVM, Texas Instruments, not shown). 
The fluidic pressure source (e.g., vacuum) is supplied off-board by either a compressor or building pressure line.

Both the LabJack and Arduino are connected to the drive computer, along with the capacitance-to-digital converter, to output PWM signals to the solenoids and read in pressure and capacitance data. The following section describes the structure of the ROS nodes for publishing, converting, and storing these data.

\subsection{ROS Nodes}
\begin{figure}[h]
    \centering
    \includegraphics[width=6.5in]{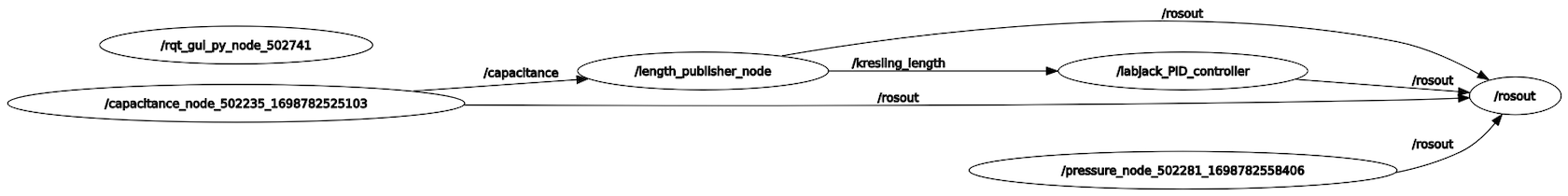}
    \caption{A ROS graph of the nodes within the feedback control loop.}
    \label{fig:ROSgraph}
\end{figure}

The PWM values were set by custom ROS \cite{quigley2009ros} nodes that publish capacitance and calculate length.
The major nodes and functions are described below and a ROS graph is depicted in Fig \ref{fig:ROSgraph}.

\begin{itemize}
    \item \textbf{capacitance\_node}: The {capacitance\_node} oversees the capacitance data measurement. It receives data from the electrodes and publishes the capacitance data of all three electrodes on three separate channels on the \textbf{\/capacitance} topic.
    \item \textbf{length\_publisher\_node}: This node acts as both a subscriber and a publisher. It subscribes to the \textbf{\/capacitance} topic, filters the data and then converts the measurements into an actuator length estimate. The length is then published on the \textbf{\/kresling\_length} topic.
    \item \textbf{labjack\_PID\_controller}: This node implements the control loop logic. It subscribes to the \textbf{\/kresling\_length} topic and uses the calculated length to set the output pwm based on the implemented logic for the given application. The ROS package included in the github repository contains several of these nodes for open loop and closed loop implementations as well as single and multi-kresling applications.
    \item \textbf{pressure\_node}: The pressure node reads the measured pressure, converts it to gauge pressure and publishes the gauge pressure on the \textbf{pressure} topic. Although pressure measurements are not currently utilized within the control mechanisms, the function is enabled to support future investigations. 
\end{itemize}

\subsection{Closed-loop Control}
\begin{figure}[h]
    \centering
    \includegraphics[width=0.7\linewidth]{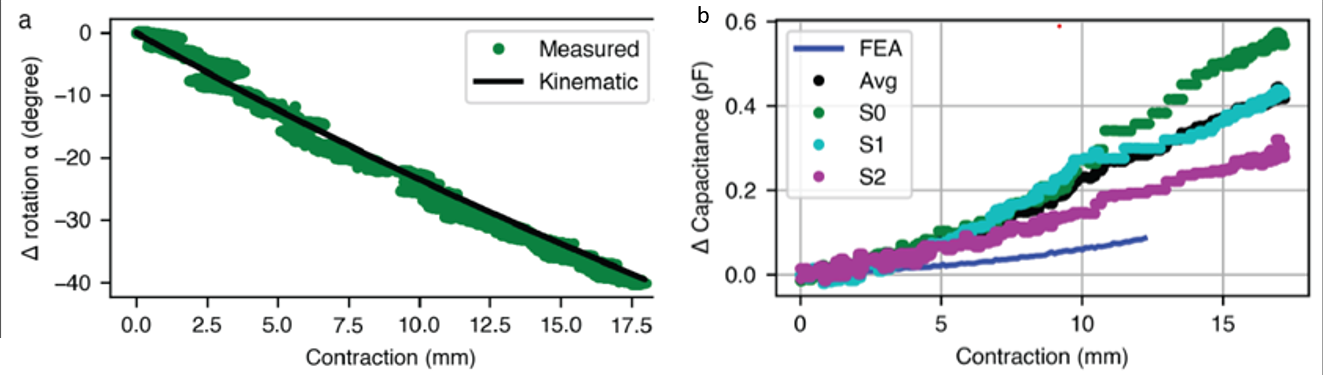}
    \caption{Open-loop response with contraction. (a) Measured and kinematic model of top face rotation $\alpha$. (b) FEA and measured capacitance change, which is zero at the fully extended length. }
    \label{fig:Capacitance FEA}
\end{figure}

The kinematic model was combined with the FEA model of capacitance ($C_{FEA}$) with $\xi$ to estimate the relationship between $L$ and capacitance (Fig. \ref{fig:Capacitance FEA}a).
This relationship was represented in the length calculation function as a third-order polynomial fit. 
Because humidity, local electromagnetic fields, and small variations in electrode placement affect robot capacitance and the match to the FEA model, we mapped $C_{FEA}$ to the measured capacitance as $ C_{meas} = \kappa^{-1}C_{FEA}$, where $\kappa$ is a factor that scales $ C_{meas}$.

After attempting multiple control approaches, including proportional-integral-derivative (PID) and discrete-time control with extensive parameter tuning, we selected discrete-time control due to its high stability and performance in this system. 
The value of the PWM signal at time $t$ that was fed to the solenoid valves on the fluidic control board was 
\begin{equation}
\text{PWM}_t = \text{PWM}_{t-1} + K_p\Delta L + K_d\frac{\Delta L}{\Delta t}
\label{eqn:pwm}
\end{equation}

where PWM$_{t-1}$ is the PWM value of the previous time step, $K_p$ is the proportional gain constant, $K_d$ is the derivative gain constant, $\Delta L$ is the error between the setpoint and the measured contraction, and $\Delta t$ is the time step.
The proportional and derivative gain values were tuned by hand with values of 0.005 and 0.001, respectively. 
The center of mass position and rotation of the actuator were measured using a commercial motion capture system (V120:Trio, OptiTrack) and served as the ground truth. %

\section{Case Study: Proprioceptive Feedback Control}
\begin{figure}[H]
    \centering
    \includegraphics[width = 3.25in]{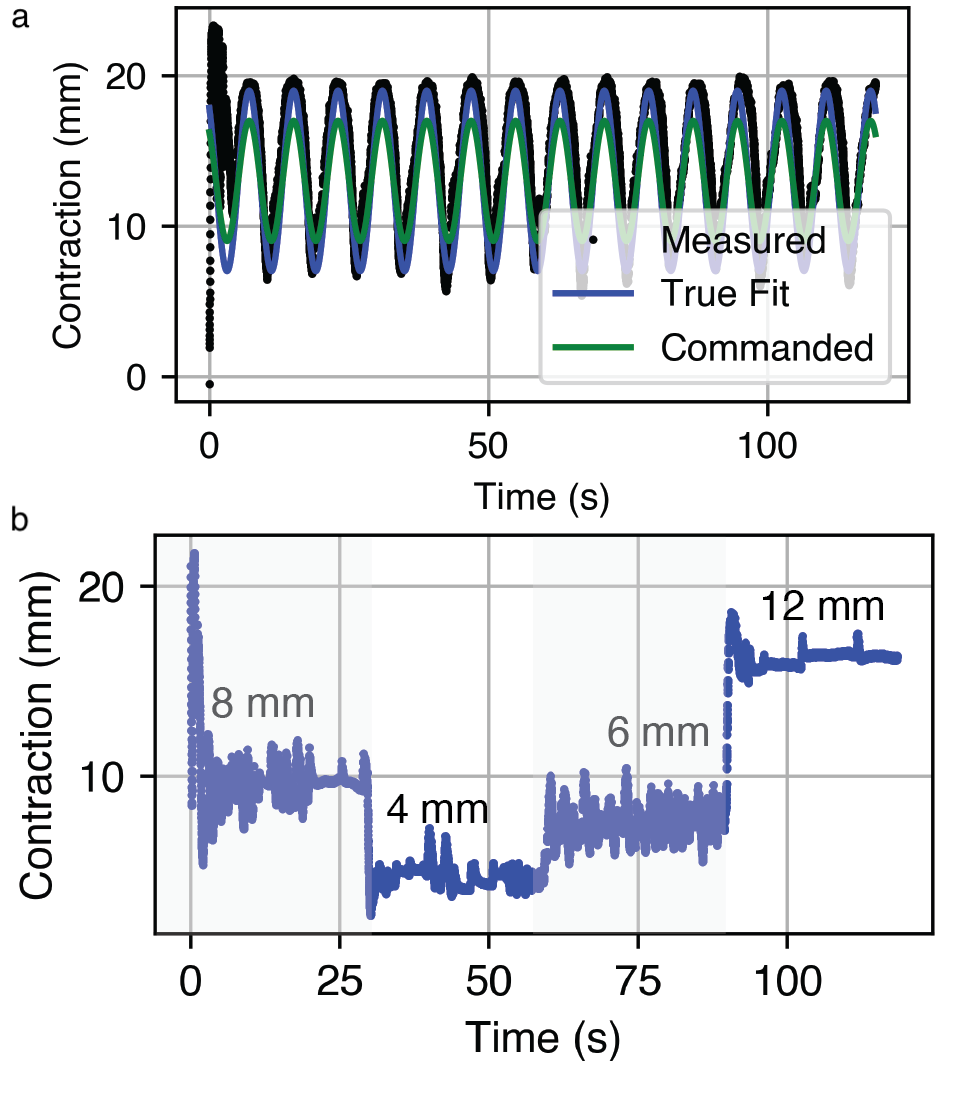}
    \caption{Feedback control response.
    (a) The response to a commanded sinusoidal input and (b) multiple control input steps.}
    \label{fig:feedback}
\end{figure}

Proprioceptive feedback control will be an essential capacity in soft robots to make them portable and used in applications where vision-based tracking is not viable. The response of the Kresling robot to a time-varying sinusoidal and stepped control setpoints is shown in Fig. \ref{fig:feedback}. RMSE is below 4 mm for setpoint contractions up to 18 mm.
The robot uses the change in capacitance due to displacement to transition between setpoints and track a commanded sine signal.
The video for this case study is available at \url{https://parses.sites.northeastern.edu/files/2023/10/Kresling_RAL.mp4}.

\section{Case Study: Origami Balance Board}
The assembled two-dimensional ``Origami Balance Board'' can serve as a lightweight and portable platform that is capable of supporting the weight of a person. 
This can be utilized in exercise and physiotherapy routines, potentially aiding in balance training, muscle strengthening and refining fine motor skills. 
Similar commercialized boards are passive and cannot adjust to the needs of a user as they build strength or improve balance through use.

Fig. \ref{fig:balance} illustrates the ability of the Kresling actuators in this envisioned application. The commanded position of each Kresling is randomly modified by decreasing the internal pressure supplied to the actuator below the working pressure of $\approx$ 35 psi. 
Under the rider's weight, the actuator then contracts and the top platform tilts. 
Supplying increased pressure again raises the rider to full, level height. The video for this case study is available at: \url{https://parses.sites.northeastern.edu/files/2023/11/Case_balance.mp4}  

\begin{figure}[h]
    \centering
    \includegraphics[width=6.5in]{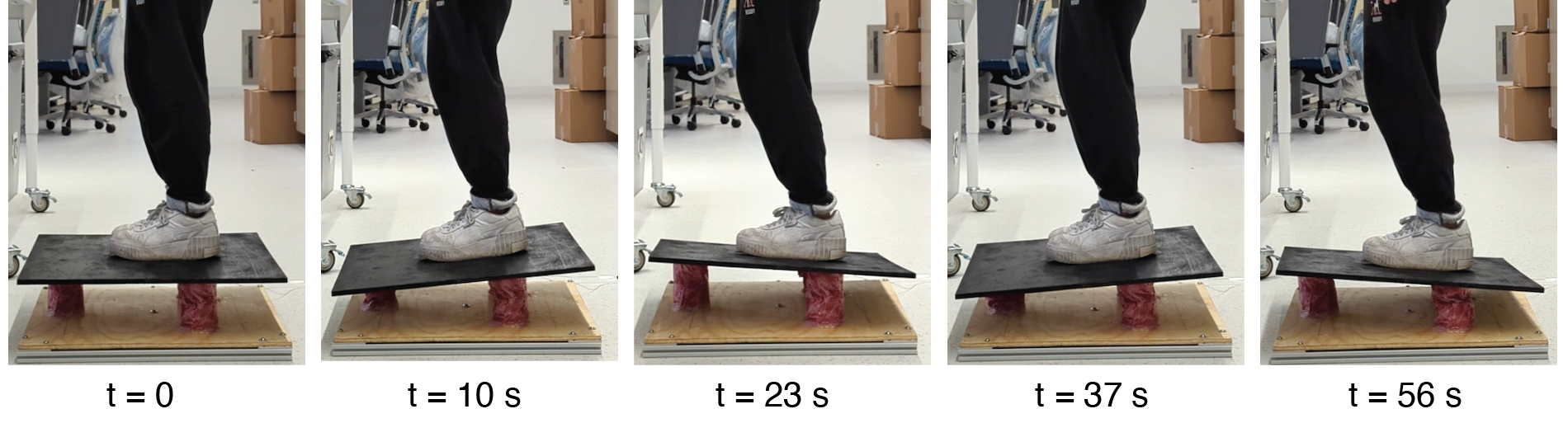}
    \caption{Photographs illustrating the motion of the balance board and rider}
    \label{fig:balance}
\end{figure}

\section{Contribution Significance}
%Please include the significance of your project to the Soft Robotics field. How will this contribution progress the field in a meaningful way?

Our project demonstrates soft, 3D-printed origami actuators with demonstrated proprioceptive feedback control or the ability to support human-scale loads.
In one case study, we demonstrated feedback position control without visual tracking, using only capacitive electrodes and discrete-time control. 
These allowed the actuator to reach setpoint states and track sinusoidal trajectories. 
In the second case study, we demonstrated a practical application of an active balance board, which required three Kresling actuators (total mass of 183.6 g). 
These actuators were able to support and manipulate a rider of approximately 60 kg.
We anticipate demonstrating the ability simultaneous support large loads and proprioceptively control actuator position in future work.

Integrating soft robots into daily life will require two types of translation: translation of real world needs into soft robot capabilities, which we demonstrated through the case studies, and translation of soft robot designs between researchers to permit deeper research investigations into soft actuator phenomena and capabilities.
Through this design document, we detailed our progress in designing soft, origami-inspired actuators, including publicly available code to generate the origami structures; a revised kinematic model that builds on previous work in origami actuators; an optimized design for proprioceptive capacitive sensors; a fabrication process that requires only an FDM printer, computer controlled cutter, and benchtop space; designs for a control board capable of positive or negative fluidic pressure; and a set of ROS nodes for controlling actuator position. 
We anticipate that this documentation will permit other soft robotics researchers to use the basic origami actuator structure for their own investigations and applications.

\pagebreak
\bibliographystyle{IEEEtran}
\bibliography{references}

\begin{thebibliography}{10}
\providecommand{\url}[1]{#1}
\csname url@rmstyle\endcsname
\providecommand{\newblock}{\relax}
\providecommand{\bibinfo}[2]{#2}
\providecommand\BIBentrySTDinterwordspacing{\spaceskip=0pt\relax}
\providecommand\BIBentryALTinterwordstretchfactor{4}
\providecommand\BIBentryALTinterwordspacing{\spaceskip=\fontdimen2\font plus
\BIBentryALTinterwordstretchfactor\fontdimen3\font minus \fontdimen4\font\relax}
\providecommand\BIBforeignlanguage[2]{{%
\expandafter\ifx\csname l@#1\endcsname\relax
\typeout{** WARNING: IEEEtran.bst: No hyphenation pattern has been}%
\typeout{** loaded for the language `#1'. Using the pattern for}%
\typeout{** the default language instead.}%
\else
\language=\csname l@#1\endcsname
\fi
#2}}

\bibitem{robertson2021soft}
M.~A. Robertson, O.~C. Kara, and J.~Paik, ``Soft pneumatic actuator-driven origami-inspired modular robotic “pneumagami”,'' \emph{The International Journal of Robotics Research}, vol.~40, no.~1, pp. 72--85, 2021.

\bibitem{wu_stretchable_2021}
S.~Wu, Q.~Ze, J.~Dai, N.~Udipi, G.~H. Paulino, and R.~Zhao, ``Stretchable origami robotic arm with omnidirectional bending and twisting,'' \emph{Proceedings of the National Academy of Sciences}, vol. 118, no.~36, p. e2110023118, 2021.

\bibitem{li_vacuum-driven_2019}
S.~Li, J.~J. Stampfli, H.~J. Xu, E.~Malkin, E.~V. Diaz, D.~Rus, and R.~J. Wood, ``A vacuum-driven origami “magic-ball” soft gripper,'' in \emph{2019 International Conference on Robotics and Automation (ICRA)}.\hskip 1em plus 0.5em minus 0.4em\relax IEEE, 2019, pp. 7401--7408.

\bibitem{park_reconfigurable_2022}
Y.~Park, J.~Kang, and Y.~Na, ``Reconfigurable shape morphing with origami-inspired pneumatic blocks,'' \emph{IEEE Robotics and Automation Letters}, vol.~7, no.~4, pp. 9453--9460, 2022.

\bibitem{chen_origami_2022}
Q.~Chen, F.~Feng, P.~Lv, and H.~Duan, ``Origami spring-inspired shape morphing for flexible robotics,'' \emph{Soft Robotics}, vol.~9, no.~4, pp. 798--806, 2022.

\bibitem{kresling_origami-structures_2012}
B.~Kresling, ``Origami-structures in nature: lessons in designing “smart” materials,'' \emph{MRS Online Proceedings Library (OPL)}, vol. 1420, pp. mrsf11--1420, 2012.

\bibitem{kim_3d_2022}
T.-H. Kim, C.~Bao, Z.~Chen, and W.~S. Kim, ``3d printed leech-inspired origami dry electrodes for electrophysiology sensing robots,'' \emph{NPJ Flexible Electronics}, vol.~6, no.~1, p.~5, 2022.

\bibitem{murali_babu_programmable_2023}
S.~P.~M. Babu, R.~Das, B.~Mazzolai, and A.~Rafsanjani, ``Programmable inflatable origami,'' in \emph{2023 IEEE International Conference on Soft Robotics (RoboSoft)}.\hskip 1em plus 0.5em minus 0.4em\relax IEEE, 2023, pp. 1--6.

\bibitem{ze_soft_2022}
Q.~Ze, S.~Wu, J.~Nishikawa, J.~Dai, Y.~Sun, S.~Leanza, C.~Zemelka, L.~S. Novelino, G.~H. Paulino, and R.~R. Zhao, ``Soft robotic origami crawler,'' \emph{Science advances}, vol.~8, no.~13, p. eabm7834, 2022.

\bibitem{kaufmann_harnessing_2021}
J.~Kaufmann, P.~Bhovad, and S.~Li, ``Harnessing the multistability of kresling origami for reconfigurable articulation in soft robotic arms,'' \emph{Soft Robotics}, vol.~9, no.~2, pp. 212--223, 2022.

\bibitem{best_new_2016}
C.~M. Best, M.~T. Gillespie, P.~Hyatt, L.~Rupert, V.~Sherrod, and M.~D. Killpack, ``A new soft robot control method: Using model predictive control for a pneumatically actuated humanoid,'' \emph{IEEE Robotics \& Automation Magazine}, vol.~23, no.~3, pp. 75--84, 2016.

\bibitem{tapia_makesense_2020}
J.~Tapia, E.~Knoop, M.~Mutn{\`y}, M.~A. Otaduy, and M.~B{\"a}cher, ``Makesense: Automated sensor design for proprioceptive soft robots,'' \emph{Soft robotics}, vol.~7, no.~3, pp. 332--345, 2020.

\bibitem{della_santina_dynamic_2018}
C.~Della~Santina, R.~K. Katzschmann, A.~Biechi, and D.~Rus, ``Dynamic control of soft robots interacting with the environment,'' in \emph{2018 {IEEE} {International} {Conference} on {Soft} {Robotics} ({RoboSoft})}, Apr. 2018, pp. 46--53.

\bibitem{patterson_untethered_2020}
Z.~J. Patterson, A.~P. Sabelhaus, K.~Chin, T.~Hellebrekers, and C.~Majidi, ``An {Untethered} {Brittle} {Star}-{Inspired} {Soft} {Robot} for {Closed}-{Loop} {Underwater} {Locomotion},'' in \emph{2020 {IEEE}/{RSJ} {International} {Conference} on {Intelligent} {Robots} and {Systems} ({IROS})}, Oct. 2020, pp. 8758--8764, iSSN: 2153-0866.

\bibitem{kim20213d}
T.-H. Kim, J.~Vanloo, and W.~S. Kim, ``3d origami sensing robots for cooperative healthcare monitoring,'' \emph{Advanced Materials Technologies}, vol.~6, no.~3, p. 2000938, 2021.

\bibitem{sun_repeated_2022}
Y.~Sun, J.~Wang, and C.~Sung, ``Repeated jumping with the rebound: Self-righting jumping robot leveraging bistable origami-inspired design,'' in \emph{2022 International Conference on Robotics and Automation (ICRA)}.\hskip 1em plus 0.5em minus 0.4em\relax IEEE, 2022, pp. 7189--7195.

\bibitem{wang_stick_2022}
Z.~Wang, H.~Wu, Z.~Feng, and H.~Wang, ``A stick on, film-like, split angle sensor via magnetic induction for versatile applications,'' \emph{IEEE Transactions on Instrumentation and Measurement}, vol.~71, pp. 1--9, 2022.

\bibitem{yan_origami-based_2023}
W.~Yan, S.~Li, M.~Deguchi, Z.~Zheng, D.~Rus, and A.~Mehta, ``Origami-based integration of robots that sense, decide, and respond,'' \emph{Nature Communications}, vol.~14, no.~1, p. 1553, 2023.

\bibitem{sun_origami-inspired_2022}
Y.~Sun, D.~Li, M.~Wu, Y.~Yang, J.~Su, T.~Wong, K.~Xu, Y.~Li, L.~Li, X.~Yu, \emph{et~al.}, ``Origami-inspired folding assembly of dielectric elastomers for programmable soft robots,'' \emph{Microsystems \& Nanoengineering}, vol.~8, no.~1, p.~37, 2022.

\bibitem{hanson_controlling_2023}
N.~Hanson, S.~F. Roberts, I.~A. Mensah, C.~Wu, J.~Healey, J.~Donelle~Furline, and K.~L. Dorsey, ``Controlling the fold: Proprioceptive feedback in a soft origami robot,'' \emph{under review}, 2023.

\bibitem{bhovad_peristaltic_2019}
P.~Bhovad, J.~Kaufmann, and S.~Li, ``Peristaltic locomotion without digital controllers: Exploiting multi-stability in origami to coordinate robotic motion,'' \emph{Extreme Mechanics Letters}, vol.~32, p. 100552, 2019.

\bibitem{lagarias1998convergence}
J.~C. Lagarias, J.~A. Reeds, M.~H. Wright, and P.~E. Wright, ``Convergence properties of the nelder--mead simplex method in low dimensions,'' \emph{SIAM Journal on optimization}, vol.~9, no.~1, pp. 112--147, 1998.

\bibitem{quigley2009ros}
M.~Quigley, K.~Conley, B.~Gerkey, J.~Faust, T.~Foote, J.~Leibs, R.~Wheeler, A.~Y. Ng, \emph{et~al.}, ``Ros: an open-source robot operating system,'' in \emph{ICRA workshop on open source software}, vol.~3, no. 3.2.\hskip 1em plus 0.5em minus 0.4em\relax Kobe, Japan, 2009, p.~5.

\end{thebibliography}
%Please include a Bibliography of works cited as part of your project
%If this is a work that is already published, or on track to be published, please include this as a citation or any relevant information in this section.

%Demonstration Video
%Detailed photo/s of your design showcasing its use case
%This must include engineering designs such as CAD, 3D modeling, circuit designs, etc. delineating your project and the way it works.

\end{document}